\documentclass[letterpaper, 10 pt, conference]{ieeeconf}  

\IEEEoverridecommandlockouts                              

\usepackage{booktabs} 
\usepackage{lipsum}   
\usepackage{adjustbox}
\usepackage{multirow}
\usepackage{amsmath} 
\usepackage{amssymb}
\usepackage{pifont}
\usepackage{xcolor}

\title{\LARGE \bf
CoDynTrust: Robust Asynchronous Collaborative Perception via Dynamic Feature Trust Modulus
}

\author{Yunjiang Xu$^{1}$, Lingzhi Li$^{1*}$, Jin Wang$^{2*}$, Benyuan Yang$^{2}$, Zhiwen Wu$^{1}$, Xinhong Chen$^{3}$, Jianping Wang$^{3}$
\thanks{$^{1}$School of Computer Science and Technology, Soochow University, Suzhou 215006, China.}%
\thanks{$^{2}$School of Future Science and Engineering, Soochow University, Suzhou 215299, China.}%
\thanks{$^{3}$Computer Science of Department, City University of Hong Kong.}%
\thanks{Emails: {\tt\small yjxu95@stu.suda.edu.cn, \newline \{lilingzhi, wjin1985, byyang\}@suda.edu.cn, zwwu@stu.suda.edu.cn, \{xinhong.chen, jianwang\}@cityu.edu.hk}}%
\thanks{$^{*}$Corresponding author.}%
}

\begin{document}

\maketitle
\thispagestyle{empty}
\pagestyle{empty}

\begin{abstract}

Collaborative perception, fusing information from multiple agents, can extend perception range so as to improve perception performance. However, temporal asynchrony in real-world environments, caused by communication delays, clock misalignment, or sampling configuration differences, can lead to information mismatches. If this is not well handled, then the collaborative performance is patchy, and what's worse safety accidents may occur. To tackle this challenge, we propose CoDynTrust, an uncertainty-encoded asynchronous fusion perception framework that is robust to the information mismatches caused by temporal asynchrony. CoDynTrust generates dynamic feature trust modulus (DFTM) for each region of interest by modeling aleatoric and epistemic uncertainty as well as selectively suppressing or retaining single-vehicle features, thereby mitigating information mismatches. We then design a multi-scale fusion module to handle multi-scale feature maps processed by DFTM. Compared to existing works that also consider asynchronous collaborative perception, CoDynTrust combats various low-quality information in temporally asynchronous scenarios and allows uncertainty to be propagated to downstream tasks such as planning and control. Experimental results demonstrate that CoDynTrust significantly reduces performance degradation caused by temporal asynchrony across multiple datasets, achieving state-of-the-art detection performance even with temporal asynchrony. The code is available at https://github.com/CrazyShout/CoDynTrust.

\end{abstract}

\section{INTRODUCTION}

With the rapid development of autonomous driving, 3D object detection, a core technology for environmental perception, has gained widespread attention \cite{3d:pointpillars, 3d:chen20203d, 3d:SAFDNet}. Its goal is to accurately identify and locate objects using sensor data from LiDAR and cameras. However, single-vehicle sensors have limited coverage, especially in complex urban environments with occlusion and reduced visibility \cite{3d:tc_transformer, 3d:occlusion}. This makes comprehensive perception difficult. As an alternative, collaborative perception arises, which can significantly improve perception range and accuracy by sharing and processing data among multiple agents via vehicle-to-everything (V2X) technology \cite{cp:quest, cp:cmimc, cp:scope}.

\begin{figure}[htbp]
  \centering
  \includegraphics[width=0.8\columnwidth, trim=30pt 25pt 30pt 15pt, clip]{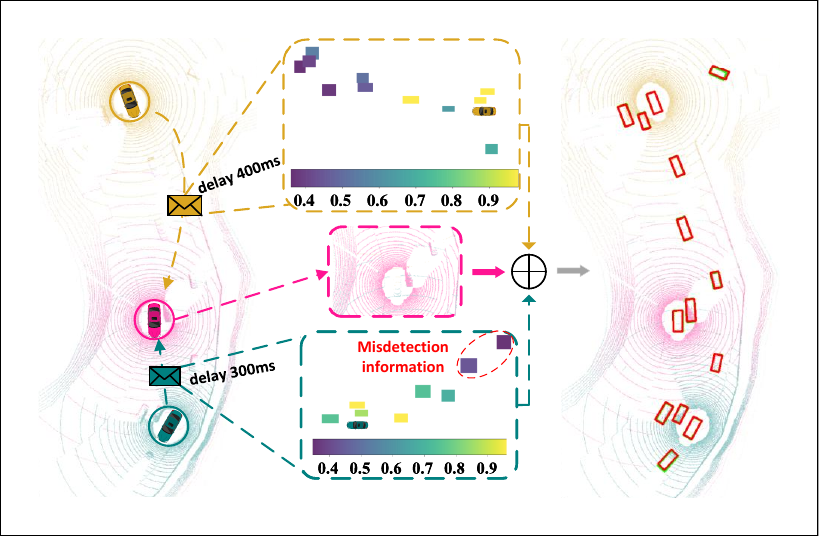}
  \caption{Based on DFTM, CoDynTrust can mitigate error propagation amplified by temporal asynchrony, effectively suppressing low-quality information and enhancing detection robustness.}
  \label{fig_head}
\end{figure}

Although collaborative perception mitigates the limitations of single-vehicle perception, it still faces practical issues such as positioning errors \cite{cp:coalign}, bandwidth limitations \cite{cp:where2comm}, and latency \cite{cp:SyncNet}. Among them, temporal asynchrony is one of the most difficult challenges since it can lead to mismatches in sensor data, thus causing collisions and conflicts between different vehicles. There exist several methods that use delay compensation strategies to address temporal asynchrony \cite{cp:SyncNet, cp:v2vnet, cp:v2xvit, cp:FFNet, cp:cobevflow}. However, they are limited to processing single-frame data and do not fully utilize historical information. Moreover, their performance is limited in high-latency or complex scenarios. In reality, asynchronous communication is inevitable due to unsynchronized sampling, different processing speed, or network congestion. This poses threats to the accuracy and real-time performance of perception.

To address the temporal asynchrony in collaborative perception, we propose CoDynTrust, as illustrated in Fig.~\ref{fig_head}. CoDynTrust evaluates the dynamic feature trust modulus (DFTM) of each region of interest (ROI) so as to tackle error propagation in the ROI generator. Furthermore, we adopt a simple linear extrapolation method instead of a learned motion prediction model so as to ensure reasonable vehicle movement within common latency ranges (0s--0.5s). Then, we design an adaptive fusion method to enhance system performance. Finally, we conduct extensive experiments on real and simulated datasets for LiDAR-based 3D object detection tasks, including DAIR-V2X \cite{cp:Dair-v2x}, V2XSet \cite{cp:v2xvit}, and OPV2V \cite{cp:OPV2V}. Experimental results show that: i) DFTM effectively addresses detection quality inconsistencies caused by noise, model flaws, and asynchronous delays. ii) CoDynTrust consistently outperforms previous state-of-the-art methods in delay scenarios. In summary, the main contributions of this paper are as follows:  
\begin{itemize}
\item We propose CoDynTrust, a novel LiDAR-based 3D detection framework for robust asynchronous multi-agent perception, addressing detection errors from noise, model defects, and worsened by delays.
\item We introduce DFTM, an uncertainty-based method estimating each agent's ROI trustworthiness. Combined with adaptive fusion, DFTM adjusts feature fusion and can be propagated to downstream tasks (e.g., planning, control) to enhance system robustness.  
\item We validate CoDynTrust via extensive experiments. Results show that CoDynTrust significantly mitigates the performance degradation caused by temporal asynchrony and improves robustness in delay scenarios.

\end{itemize}

\section{RELATED WORKS}

\subsection{Collaborative Perception}
Collaborative perception addresses the challenges of limited perception range and occlusion in single-vehicle autonomous driving by enabling information exchange between agents. High-quality datasets like DAIR-V2X \cite{cp:Dair-v2x}, V2XSet \cite{cp:v2xvit}, OPV2V \cite{cp:OPV2V}, and V2X-Sim \cite{cp:v2x-sim} have be given, and notable methods are proposed, such as DiscoNet \cite{cp:disconet} using a teacher-student distillation framework to extract more information during training, V2X-ViT \cite{cp:v2xvit} the first heterogeneous Transformer for V2X perception, CoAlign \cite{cp:coalign} which enhances consistency between agents using pose graph modeling, and CoBEVFlow \cite{cp:cobevflow} which mitigates delays through motion prediction.


\subsection{Temporal Asynchrony Issue in Collaborative Perception}


Temporal asynchrony is unavoidable in real-world environments, exacerbating errors caused by noise and model deficiencies. Existing solutions like V2VNet \cite{cp:v2vnet} employ CNNs for delay compensation, while V2X-ViT \cite{cp:v2xvit} mitigates spatio-temporal distortions via position encoding. However, these methods insufficiently leverage historical data. Though SyncNet \cite{cp:SyncNet} applies Conv-LSTM \cite{common:conv-lstm} for delay handling, its RNN-based feature generation struggles with noise and irregular delays. FFNet \cite{cp:FFNet} processes irregular multi-frame data but introduces generation noise and infrastructure dependencies. CoBEVFlow \cite{cp:cobevflow} demands high-quality ROI generation, and its three-stage training risks error propagation from delays. To overcome these limitations, we propose an optimized asynchronous collaborative perception framework.


\subsection{Uncertainty Quantiﬁcation}

Uncertainty quantification critically evaluates prediction reliability through two categories: aleatoric (data noise irreducible with data volume) and epistemic (reducible model limitations). Aleatoric uncertainty is captured through direct modeling (DM) \cite{unc:dm} with variance prediction outputs; epistemic uncertainty estimated via Monte Carlo dropout \cite{unc:mcdropout} approximating Bayesian inference or deep ensembles \cite{unc:deepensemble}. In autonomous driving, this enhances safety in perception \cite{unc:UMoE}, \cite{unc:multi-modal_DiFeng} and decision-making. Current collaborative perception research predominantly applies uncertainty to downstream tasks like planning and control \cite{unc:doubleM}, \cite{cp:mot-cup}, neglecting specific challenge resolution. In CoDynTrust, we quantify both uncertainties using DM and MC dropout to guide asynchronous collaborative perception, filtering out noise and low-quality data, and propagating uncertainty to tasks such as planning and control.

\section{PROBLEM FORMULATION}

Consider a scenario with $N$ agents, where each agent can exchange information with others and store two frames of historical data from others. For the $n$-th agent, let $\mathbf{X}_n^{t_n^i}$ and $\mathbf{Y}_n^{t_n^i}$ respectively represent the perception observation and supervision at time $t_n^i$. Here, $t_n^i$ is the timestamp of agent $n$ at the $i$-th moment. $\mathcal{P}_{m \rightarrow n}^{t_m^j}$ denotes the collaborative message sent from agent $m$ to agent $n$ at $t_m^j$. Temporal asynchrony includes: i) timestamp misalignment between agents, $t_{m}^{j} \ne t_{n}^{i}$, and ii) irregular time intervals between two frames of received messages from other agents, i.e., $t_{m}^{j} - t_{m}^{j-1}$. Therefore, the task of asynchronous collaborative perception can be formulated as:
\begin{align*}
&\underset{\theta, \mathcal{P}}{\arg \max } \sum_{n=1}^{N} g\left(\hat{\mathbf{Y}}_{n}^{t_n^i}, \mathbf{Y}_{n}^{t_n^i}\right), \tag{1}\\
\text{subject to} \quad &\hat{\mathbf{Y}}_{n}^{t_n^i} = \Phi_{\theta}\left(\mathbf{X}_{n}^{t_n^i},\left\{\mathcal{P}_{m \rightarrow n}^{t_m^{j}}, \mathcal{P}_{m \rightarrow n}^{t_m^{j-1}} \right\}_{m=1}^{N}\right) ,
\end{align*}
In this formula, $g(\cdot, \cdot)$ represents the perception evaluation metric, and $\hat{\mathbf{Y}}_{n}^{t_n^i}$ denotes the $n$-th agent's detection result at time $t_n^i$. The collaborative perception network $\Phi$ is parameterized by $\theta$. When timestamps are synchronized and frame intervals are uniform (i.e., $t_{m}^{j} = t_{n}^{i}$ and $t_{n}^{i} - t_{n}^{i-1}$ is constant), the problem simplifies to the ideal delay-free case of uniform. If $t_{m}^{j} \neq t_{n}^{i}$ but intervals remain uniform, the setup resembles SyncNet's scenario.

\begin{figure*}[t]
  \centering
  \includegraphics[width=\textwidth, trim=7.6cm 0.86cm 7.5cm 0.6cm, clip]{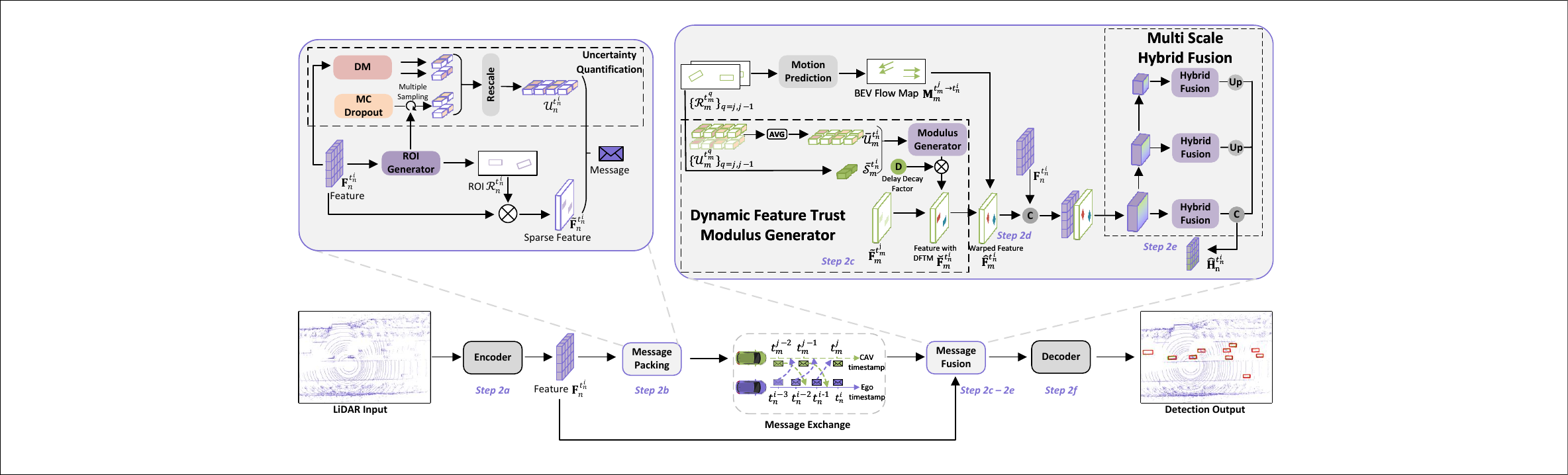}
  \caption{System overview. The message packing process prepares ROI, uncertainty, and sparse features for efficient communication and BEV flow map generation. Message fusion generate Dynamic Feature Trust Modulus and scatters it back to the sparse feature map, while the BEV map is generated and used for motion compensation. Finally, multi-scale Hybrid Fusion is applied to fuse feature maps from all agents.}
  \label{overview}
\end{figure*}


\section{ASYNCHRONY-ROBUST COLLABORATIVE Perception}


\subsection{Overall Architecture}

Temporal asynchrony means that agents perceive the same object at different times. Existing methods mitigate them by: i) using a pre-trained ROI generator after feature extraction, ii) feeding multi-frame ROIs into a motion prediction module to generate birds' eye view (BEV) flow map, and iii) warping asynchronous feature with the BEV flow map before fusion. However, such strategy has two issues: i) over-reliance on the ROI generator, leading to variable ROI quality due to noise and model limitations; and ii) errors accumulate in both ROI and BEV flow generation, worsened by irregular delays. CoDynTrust addresses these issues by: i) parallel uncertainty quantification for each ROI, and ii) using linear extrapolation for motion prediction and averaging uncertainty across frames to generate DFTM. By doing so, we can filter noise and retain high-quality features. Furthermore, an adaptive fusion further enhances feature integration. Formally, CoDynTrust can be expressed as:
\begin{align}
\mathbf{F}_n^{t_n^i} &= f_{\text{enc}}(\mathbf{X}_n^{t_n^i}), \tag{2a} \\
\mathcal{R}_n^{t_n^i}, \mathcal{U}_n^{t_n^i}, \tilde{\mathbf{F}}_n^{t_n^i} &= f_{\text{roi\_gen}}(\mathbf{F}_n^{t_n^i}), \tag{2b} \\
\{\mathbf{M}_{m}, \mathbf{D}_{m}\}^{\Delta t} &= f_{\text{flow\_gen}}(t_n^{i}, \mathbf{R}_m, \mathbf{U}_m), \tag{2c} \\
\hat{\mathbf{F}}_m^{t_n^i} &= f_{\text{warp\_dtfm}}(\tilde{\mathbf{F}}_m^{t_m^j}, \{\mathbf{M}_{m}, \mathbf{D}_{m}\}^{\Delta t}), \tag{2d} \\
\hat{\mathbf{H}}_n^{t_n^i} &= f_{\text{agg}}(\tilde{\mathbf{F}}_n^{t_n^i}, \{\hat{\mathbf{F}}_m^{t_n^i}\}_{m \in \mathcal{N}_n}), \tag{2e} \\
\hat{\mathbf{Y}}_n^{t_n^i} &= f_{\text{dec}}(\hat{\mathbf{H}}_n^{t_n^i}), \tag{2f}
\end{align}
Here, $\mathbf{F}_n^{t_n^i} \in \mathbb{R}^{H \times W \times D}$ is $n$-th agent‘s BEV feature at timestamp $t_{n}^{i}$, where $H$ and $W$ are its size and $D$ is the channels. $\mathcal{R}_n^{t_n^i}$ is the set of ROIs, $\mathcal{U}_n^{t_n^i}$ is the corresponding uncertainty, and $\tilde{\mathbf{F}}_n^{t_n^i}$ is the sparse version of $\mathbf{F}_n^{t_n^i}$ containing features within $\mathcal{R}_n^{t_n^i}$. $\{\mathbf{M}_{m}, \mathbf{D}_{m}\}^{\Delta t}$ refers to the BEV flow map $\mathbf{M}_{m}^{t_m^j\rightarrow t_n^i} \in \mathbb{R}^{H \times W \times 2}$, reflecting grid cells movement from $t_m^j$ to $t_n^i$. $\mathbf{D}_{m}^{t_m^j\rightarrow t_n^i} \in \mathbb{R}^{o\times 1}$ represents the trust modulus of each ROI, where \( o \) is the number of matched ROI pairs. $\hat{\mathbf{F}}_m^{t_n^i}$ is the re-aligned feature after motion compensation and DFTM application. $\hat{\mathbf{H}}_n^{t_n^i}$ is the aggregated feature from all agents, where $\mathcal{N}_n$ denotes the collaborative neighbors of agent $n$. $\hat{\mathbf{Y}}_n^{t_n^i}$ is the system's output.

The whole procedure of our approach is given in Fig. \ref{overview}. Specifically, Step 2a extracts BEV features from observation data. Step 2b generates ROIs and evaluates uncertainty, and then agents exchange messages including $\mathcal{U}_n^{t_n^i}$, $\mathcal{R}_n^{t_n^i}$, and $\tilde{\mathbf{F}}_n^{t_n^i}$. Step 2c generates BEV flow map. Step 2d applies the BEV flow map to warp asynchronous features and apply DFTM so as to obtain re-aligned feature map. Step 2e aggregates features from all agents. Step 2f outputs the final perception result. Note that, steps 2a and 2b run without communication, while communication begins in Steps 2c-2f. Next, we explain the main steps and contents of CoDynTrust.


\subsection{ROI Generation with Uncertainty Quantification}


\textbf{Uncertainty Quantification.} For agent $m$ at timestamp \( t_{m}^{j} \) with BEV feature \( \mathbf{F}_m^{t_m^j} \), CoDynTrust quantifies aleatoric and epistemic uncertainty. Each type of uncertainty corresponds to classification and regression uncertainty. Let \( \mathbf{P}_{\text{DM}}(\cdot) \) and \( \mathbf{P}_{\text{MCD}}(\cdot) \) represent the processes of DM and MC Dropout, respectively. The quantification is:
\begin{align}
\{u_{ale}^{cls}, u_{ale}^{reg}\}_m^{t_m^j} &= \mathbf{P}_{\text{DM}}(\mathbf{F}_m^{t_m^j}) \tag{3a} \label{eq:ale_uncertainty} \\
\{u_{epi}^{cls}, u_{epi}^{reg}\}_m^{t_m^j} &= \mathbf{P}_{\text{MCD}}(\mathbf{X}_m^{t_m^j}, \theta_{single}, T) \tag{3b} \label{eq:epi_uncertainty}
\end{align}

Since we evaluates ROI uncertainty, the regression aleatoric uncertainty is evaluated by modeling the bounding box center $(x, y)$ as Gaussian random variable. Considering the periodicity of yaw angle \( \alpha \), it is modeled \( \alpha \) with von Mises distribution. The regression head predicts the mean, and an additional head predicts the log variance (for $x$ and $y$) and the inverse concentration parameter (for \( \alpha \)) to capture uncertainty.

Similarly, the classification aleatoric uncertainty is modeled by treating the logit of the classification head as Gaussian random variable. An additional head predicts the variance to quantify classification noise. Since it is challenging to directly optimize the Gaussian parameters \cite{unc:survey}, we apply the reparameterization trick \cite{unc:vae}. Specifically, we express the random variable as a differentiable function by introducing standard normal noise, scaled and shifted for smooth gradient propagation. This approach effectively learns classification uncertainty while maintaining stability.

We use MC Dropout to quantify epistemic uncertainty. In equation (3b), \( \mathbf{X}_m^{t_m^j} \) is the raw observation of agent $m$ at timestamp \( t_{m}^{j} \), \( \theta_{single} \) is the single-vehicle detection model parameter, and $T$ is the number of MC Dropout inferences. Since dropout is used only in the decoder, during inference, step 2b is executed $T$ times. The mean of the $T$ results is taken as the final output, with classification entropy and regression variance representing the epistemic uncertainties. Finally, the raw uncertainty set for agent $m$ at \( t_{m}^{j} \) is denoted as: $(\mathcal{U}_m^{t_m^j})_{raw} = \{u_{ale}^{cls}, u_{ale}^{reg}, u_{epi}^{cls}, u_{epi}^{reg}\}_m^{t_m^j}$. Due to the differences in evaluation methods and unitless properties of uncertainty, the classification or regression uncertainties cannot be directly compared. Therefore, we perform a rescaling operation, denoted as \( \mathbf{P}_{\text{rescale}}(\cdot) \), with the process: $(\mathcal{U}_m^{t_m^j})_{dense} = \mathbf{P}_{\text{rescale}}((\mathcal{U}_m^{t_m^j})_{raw})$. Specifically, for classification uncertainty, we use the classification deviation ratio \cite{unc:UMoE} to unify the scale:
\begin{align*}
u^{cls} = \frac{\mu_u}{\mu_u +  ReLU(u_{raw}^{cls} - \mu_u - \sigma_u)} \cdot \\ \frac{\mu_s}{\mu_s + ReLU(-(s_c - \mu_s - \sigma_s))} ,  \tag{4}
\end{align*}
Here, \( u_{raw}^{cls} \) is one type of classification uncertainty, and \( s_c \) is the confidence score. \( \mu_u \), \( \sigma_u \), \( \mu_s \), and \( \sigma_s \) are the mean and variance of raw classification uncertainty and confidence scores for positive samples in the test set. Compared to raw classification uncertainty, the classification deviation ratio provides more information and amplifies the difference between positive and negative samples. Finally, regression uncertainty is scaled by multiplying the uncertainty of the bounding box center \((x, y)\) by the box's diagonal and applying Z-Score normalization for a unified scale.

\textbf{ROI Set Generation.} The BEV feature from step 2a is fed into the ROI generator to generate the ROI: $\mathbf{O}_{m}^{t_{m}^{j}} = \Phi_{roi\_gen}^{T}\left(\mathbf{F}_{m}^{t_{m}^{j}}\right) \in \mathbb{R}^{H \times W \times 8}$,
where \( \Phi_{roi\_gen}^{T}\left(\cdot\right) \) is the ROI generation network, and $T$ is the number of MC Dropout inferences. Each element \( (\mathbf{O}_{m}^{t_{m}^{j}})_{h,w} = \left( c, x, y, z, d_x, d_y, d_z,  \alpha \right) \) is the mean of the $T$ detection results for class confidence, position, size, and orientation of an ROI. Post-processing then generates detection boxes, projected into BEV space as the ROI set \( \mathcal{R}_m^{t_m^j} \), with each element \( \left( c, x, y, d_x, d_y, \alpha \right) \). Finally, the uncertainty set \( \mathcal{U}_m^{t_m^j} \) corresponding to the ROI set is indexed.

\textbf{Sparse Feature Map Generation.} A binary mask \( \mathbf{H} \in \mathbb{R}^{H \times W} \) is generated from the ROI set, where regions inside are 1 and outside are 0. Applying this mask to the feature map from step 2a, we have \( \tilde{\mathbf{F}}_m^{t_m^j} = \mathbf{F}_m^{t_m^j} \odot \mathbf{H} \), retaining only ROI features. Agent $m$ then packages \( \mathcal{U}_m^{t_m^j} \), \( \mathcal{R}_m^{t_m^j} \), and \( \tilde{\mathbf{F}}_m^{t_m^j} \), and sends them to other agents.

\subsection{Flow Generation with DFTM}

After communication, collaborative nodes receive information from others, often with irregular time delays. To address this, we require: i) efficient and interpretable motion prediction to avoid error amplification; and ii) dynamic feature trust modulus to evaluate historical data to reduce noise. Step 2c can achieve so. Specifically, a BEV flow map is generated to record displacements and calibrate the received features. Unlike learning-based models that require multiple frames, CoDynTrust only needs two. Each frame's ROIs are matched as detection proposals using an efficient algorithm \cite{cp:cobevflow}, followed by motion prediction generating the BEV flow map.

CoDynTrust uses linear extrapolation for motion prediction. For the $r$-th ROI of agent $m$ at timestamp \( t_{m}^{j} \), the motion attributes are extracted, i.e., \( \{\mathbf{P}_r^{t_m^{j}}, \mathbf{P}_r^{t_m^{j-1}} \} \), where \( \mathbf{P}_r^{t_m^{j}} = (x_r^{t_m^j}, y_r^{t_m^j}, \alpha_r^{t_m^j}) \). The ROI velocity is \( v_r = (\mathbf{P}_r^{t_m^j} - \mathbf{P}_r^{t_m^{j-1}})/(t_m^j-t_m^{j-1}) \), and the motion displacement is \( \Delta d_r = v_r  \times (t_n^i-t_m^{j}) \). \( \Delta d_r \) is then used to create the BEV flow map \( \mathbf{M}_{m}^{t_m^j\rightarrow t_n^i} \), storing motion mappings in two axes for each grid in BEV space within the ROI area, while other regions retain an identity mapping.

\begin{table*}[t] 
    \centering
    \caption{Performance Comparison of CoDynTrust and Baseline Methods at Expected Delays from 0ms to 500ms on Three Datasets.}
    \begin{adjustbox}{max width=\textwidth}
        \begin{tabular}{ccc|ccc|ccc|ccc}
            \toprule
            \multicolumn{3}{c|}{Dataset} & \multicolumn{3}{c}{DAIR-V2X} & \multicolumn{3}{c}{V2XSet} & \multicolumn{3}{c}{OPV2V(Culver)} \\ 
            \multicolumn{3}{c|}{Expectation of interval (ms)} & 0 & 300 & 500 & 0 & 300 & 500 & 0 & 300 & 500 \\  \cmidrule{1-12}
            \multicolumn{3}{c|}{Model / Metric} & \multicolumn{9}{c}{AP@0.5 $\uparrow$} \\  \cmidrule{1-12}
            \multicolumn{3}{c|}{Single} & \multicolumn{3}{c|}{0.6349} & \multicolumn{3}{c|}{0.6515} & \multicolumn{3}{c}{0.6427} \\  \cmidrule{1-12}
            \multicolumn{3}{c|}{Late Fusion} & 0.6043 & 0.5401 & 0.5045 & 0.7876 & 0.4671 & 0.4112 & 0.8049 & 0.6702 & 0.6088 \\  
            \multicolumn{3}{c|}{V2VNet \cite{cp:v2vnet}} & 0.7395 & 0.6572 & 0.6298 & 0.8691 & 0.5766 & 0.4990 & 0.8133 & 0.7332 & 0.6909 \\  
            \multicolumn{3}{c|}{V2X-ViT \cite{cp:v2xvit}} & 0.7820 & 0.7202 & 0.7012 & 0.8785 & 0.6549 & 0.5535 & 0.8435 & 0.7788 & 0.7397 \\  
            \multicolumn{3}{c|}{Where2comm \cite{cp:where2comm}} & 0.7921 & 0.7103 & 0.6861 & \textbf{0.8918} & 0.6767 & 0.6131 & \textbf{0.8695} & 0.8014 & 0.7677 \\  
            \multicolumn{3}{c|}{Where2comm+SyncNet \cite{cp:SyncNet}} & 0.7921 & 0.7324 & 0.7103 & \textbf{0.8918} & 0.8081 & 0.7773 & \textbf{0.8695} & 0.8106 & 0.7892 \\  
            \multicolumn{3}{c|}{CoBEVFlow \cite{cp:cobevflow}} & 0.7751 & 0.7419 & 0.7316 & 0.8632 & 0.8542 & 0.8388 & 0.8370 & 0.8217 & 0.8186 \\  
            \multicolumn{3}{c|}{Ours} & \textbf{0.7922} & \textbf{0.7481} & \textbf{0.7357} & 0.8780 & \textbf{0.8644} & \textbf{0.8475} & 0.8646 & \textbf{0.8566} & \textbf{0.8532} \\  \cmidrule{1-12}

            \multicolumn{3}{c|}{Model / Metric} & \multicolumn{9}{c}{AP@0.7 $\uparrow$} \\  \cmidrule{1-12}
            \multicolumn{3}{c|}{Single} & \multicolumn{3}{c|}{0.4962} & \multicolumn{3}{c|}{0.5204} & \multicolumn{3}{c}{0.5213} \\  \cmidrule{1-12}
            \multicolumn{3}{c|}{Late Fusion} & 0.3746 & 0.3226 & 0.3027 & 0.6801 & 0.3239 & 0.3152 & 0.6809 & 0.5031 & 0.4719 \\  
            \multicolumn{3}{c|}{V2VNet \cite{cp:v2vnet}} & 0.5835 & 0.4982 & 0.4945 & 0.7675 & 0.3981 & 0.3715 & 0.6398 & 0.4864 & 0.4578 \\  
            \multicolumn{3}{c|}{V2X-ViT \cite{cp:v2xvit}} & 0.6590 & 0.6046 & 0.5973 & 0.7170 & 0.4618 & 0.4179 & 0.6834 & 0.5990 & 0.5689 \\  
            \multicolumn{3}{c|}{Where2comm \cite{cp:where2comm}} & 0.6920 & 0.6066 & 0.5952 & 0.7966 & 0.5026 & 0.4854 & 0.7858 & 0.6638 & 0.6393 \\  
            \multicolumn{3}{c|}{Where2comm+SyncNet \cite{cp:SyncNet}} & 0.6920 & 0.6246 & 0.6121 & 0.7966 & 0.6720 & 0.6511 & 0.7858 & 0.7057 & 0.6906 \\  
            \multicolumn{3}{c|}{CoBEVFlow \cite{cp:cobevflow}} & 0.6627 & 0.6264 & 0.6207 & 0.8111 & 0.7638 & 0.7340 & 0.7832 & 0.7400 & 0.7315 \\  
            \multicolumn{3}{c|}{Ours} & \textbf{0.7001} & \textbf{0.6431} & \textbf{0.6368} & \textbf{0.8356} & \textbf{0.7869} & \textbf{0.7582} & \textbf{0.8194} & \textbf{0.7952} & \textbf{0.7833} \\  
            \bottomrule
        \end{tabular}
    \end{adjustbox}
    \label{comparative_experiment}
\end{table*}

Then, to dynamically generate the feature trust modulus, CoDynTrust combines the confidence and uncertainty of all ROIs from historical frames: $\mathbf{D}_{m}^{t_m^j\rightarrow t_n^i} = \Phi_{dftm} (\bar{S}_{m}^{t_n^{i}}, \bar{U}_{m}^{t_n^{i}}) \times d_f$, where \( \Phi_{dftm }\left(\cdot\right) \) is the DFTM generation network, and \( \bar{\mathcal{S}}_{m}^{t_n^{i}}\) and \(\bar{\mathcal{U}}_{m}^{t_n^{i}} \) are the averages of confidence and uncertainty, respectively. Since the previous matching filters out extreme outliers, a simple average works well. \( d_f \) is the delay decay factor, resisting noise under high delays. \( \Phi_{dftm }\left(\cdot\right) \) uses residual blocks to produce the feature trust modulus between 0 and 1. After compensating and re-estimating ROI confidence and uncertainty, a raw modulus is obtained. It's then multiplied by \( d_f = \exp(-k\times (t_n^i-t_m^{j})) \), where  \( k = 0.02 \), to further adjust for low-reliability ROIs under high delays. The final output is \( \mathbf{D}_{m}^{t_m^j\rightarrow t_n^i} \in \mathbb{R}^{o \times 1} \).

\textbf{Feature Warp Guided by DFTM.} In step 2d, the BEV flow map and DFTM are applied to the feature map. Each sparse feature grid shifts based on the flow map, while DFTM is scattered back into the feature map and scales each ROI's feature. This step is denoted as \( \mathbf{P}_{\text{scatter}}(\cdot) \), and the overall transformation is: $\hat{\mathbf{F}}_m^{t_n^i}[h,w] = \mathbf{P}_{\text{scatter}}(\mathbf{D}_{m}^{t_m^j \rightarrow t_n^i}) \odot \tilde{\mathbf{F}}_m^{t_m^j} \left[ h + \mathbf{M}_{m}^{t_m^j \rightarrow t_n^i}[h,w,0], \right. \left. w + \mathbf{M}_{m}^{t_m^j \rightarrow t_n^i}[h,w,1] \right]$. Through this transformation, the feature maps of non-ego agents achieve asynchronous compensation. 

\subsection{Hybrid Fusion}

\begin{figure}[htbp]
  \centering
  \includegraphics[width=0.9\columnwidth, trim=138pt 18pt 64pt 21pt, clip]{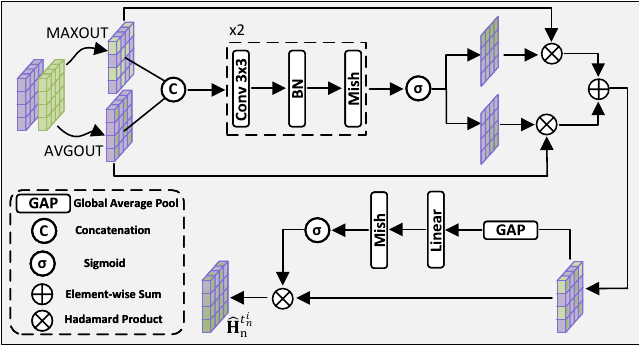}
  \caption{Overall Structure of Hybrid Fusion.}
  \label{hybrid_fusion}
\end{figure}

In step 2e, i.e., hybrid fusion as shown in Fig.~\ref{hybrid_fusion}, fuses features from other agents at multiple scales to adapt to objects of varying sizes. Since the features processed by DFTM exhibit sparsity and scaling, traditional methods struggle with effective fusion. Hybrid fusion concatenates features applying both MAXOUT \cite{cp:F-cooper} and AVGOUT operations to highlight key features and retain multi-agent perspectives. The results are concatenated and passed through convolutional layers and a sigmoid function to generate spatial weights, which are used to weight and sum the MAXOUT and AVGOUT outputs, forming the fused feature map. Global average pooling and a linear layer produce channel weights that further refine the feature map. The final output \( \hat{\mathbf{H}}_n^{t_n^i} \) integrates spatial and channel information. Furthermore, for smooth gradient propagation, we use Mish \cite{common:mish} activation throughout.

\section{EXPERIMENTAL RESULTS}

We conducted experiments on three benchmark datasets. All experiments focus on LiDAR 3D object detection.
\subsection{Dataset}
\textbf{DAIR-V2X} \cite{cp:Dair-v2x} is a real-world collaborative perception dataset with two agents (a vehicle and an infrastructure) per sample. Both capture image and point cloud data with 3D annotations. The LiDAR range is $201.6m \times 80m$ with a 10Hz sampling frequency. We applied additional annotations \cite{cp:coalign} to cover objects outside the camera’s field of view, achieving 360° detection coverage. \textbf{V2XSet} \cite{cp:v2xvit} is a V2X simulation dataset generated using Carla \cite{cp:CARLA} and OpenCDA \cite{cp:OpenCDA}, containing 73 collaborative scenarios with 11,447 frames of point cloud data. Each scenario includes 2 to 4 vehicles and infrastructure agents. The LiDAR range is $281.6m \times 80m$, with standard splits for training, validation, and testing. \textbf{OPV2V} \cite{cp:OPV2V} is a vehicle-to-vehicle collaborative perception dataset simulated by OpenCDA and Carla. It includes 70 scenarios and 11,464 frames with over 232,913 3D annotations. The Culver City Digital Twin test set, which reflects real-world conditions, is used to evaluate generalization performance.

\subsection{Impelment Details}

CoDynTrust's training has two stages: pre-training an ROI generator and training the fusion detector with frozen ROI generator parameters. Both use the same structure without sharing parameters. We use PointPillar \cite{3d:pointpillars} as the encoder with a grid size of $(0.4m, 0.4m)$. We apply Focal Loss \cite{optimize:Focalloss} for classification, weighted smooth L1 Loss for regression, and negative log-likelihood loss for uncertainty modeling. The AdamW \cite{optimize:AdamW} optimizer is used with learning rates of 0.001 for the ROI generator and 0.002 for the fusion detector.  All experiments run on an NVIDIA A100 GPU.



\subsection{Quantitative Results}

\textbf{Benchmark Comparison.} To evaluate CoDynTrust's performance under temporal asynchrony, we compared it with multiple models, both with and without asynchronous robustness designs. Referencing CoBEVFlow's setup, we applied binomially sampled frame intervals to create irregular time gaps. Table~\ref{comparative_experiment} shows the AP results at IoU thresholds of 0.5 and 0.7 on the DAIR-V2X, V2XSet, and OPV2V datasets. CoDynTrust consistently outperformed in nearly all scenarios, showing strong robustness to temporal asynchrony.

\begin{figure}[htbp]
  \centering
  \includegraphics[width=0.9\columnwidth, trim=208pt 115pt 260pt 50pt, clip]{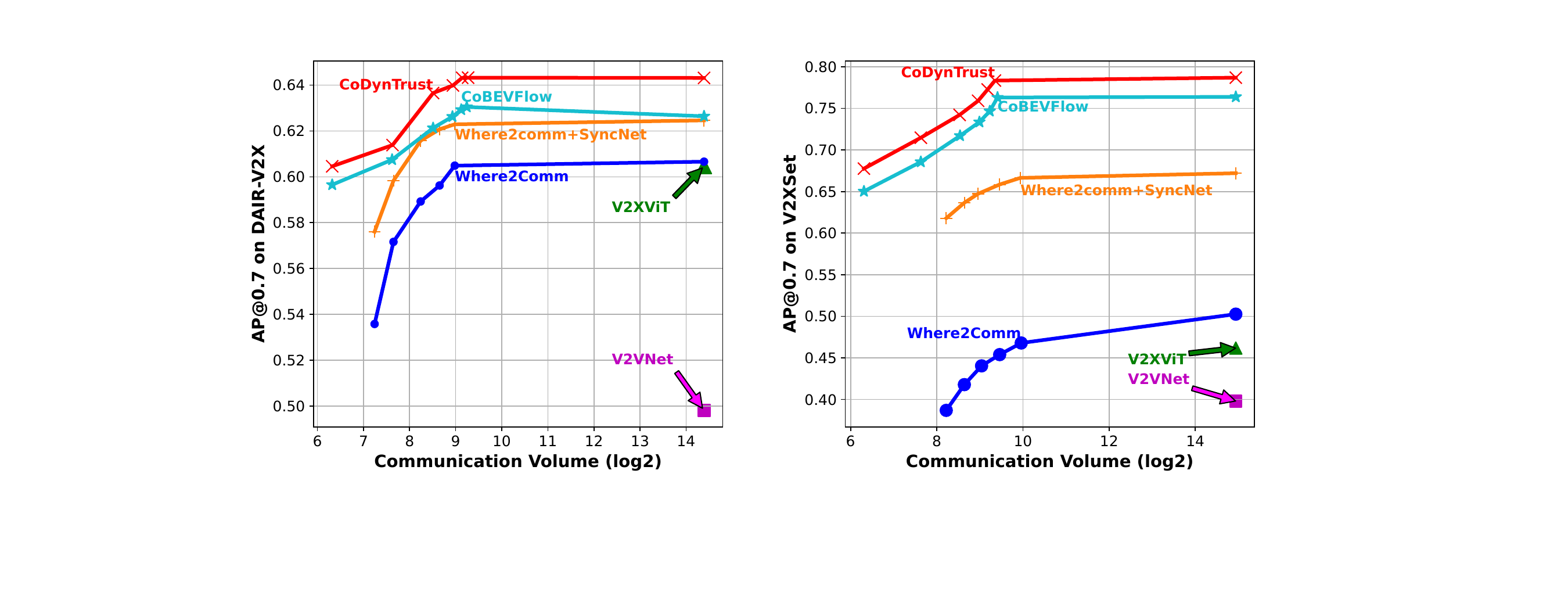}
  \caption{Trade-off between detection performance (AP@0.7) and communication bandwidth under expected 300ms delay on DAIR-V2X (left) and V2XSet (right) datasets.}
  \label{comm_volume}
\end{figure}

\textbf{Trade-off Between Detection Performance and Communication Cost.} CoDynTrust transmits uncertainty, ROIs and sparse features, resulting in low communication costs and being bandwidth-friendly. Fig.~\ref{comm_volume} shows the trade-off between bandwidth and performance (AP@0.7) under expected 300ms temporal asynchrony for CoDynTrust and other methods. Experiments on DAIR-V2X and V2XSet show: i) CoDynTrust consistently outperforms others with the same bandwidth.  
ii) On V2XSet, CoDynTrust reached its optimal performance with lower bandwidth requirements than Where2Comm.

\textbf{Robustness to Pose Errors.} To evaluate system robustness, we test performance under temporal asynchrony and pose errors. Pose errors are simulated during inference by adding Gaussian noise to the 2D center and yaw angle, with means of $0m/0^\circ$ and standard deviations ranging from $0m-0.5m$ and $0^\circ-0.5^\circ$. The expected frame interval is set to 300ms for temporal asynchrony. Table \ref{table2} compares SOTA models' robustness on the DAIR-V2X dataset with pose errors. CoDynTrust maintained top performance even with both temporal asynchrony and pose errors, outperforming CoBEVFlow by 2.86\% in AP@0.7 at $0.4m/0.4^\circ$ noise.

\begin{table}[htbp] 
    \centering
    \caption{Detection performance on the DAIR-V2X dataset with Gaussian-distributed pose noise during testing.}
    \begin{adjustbox}{max width=\columnwidth }
        \begin{tabular}{ccc|ccccc}
            \toprule
            \multicolumn{3}{c|}{Noise Level $\sigma_t/\sigma_r(m/\circ)$} & 0.0/0.0 & 0.1/0.1 & 0.2/0.2 & 0.3/0.3 & 0.4/0.4  \\  \cmidrule{1-8}
            \multicolumn{3}{c|}{Model / Metric} & \multicolumn{5}{c}{AP@0.5 $\uparrow$} \\  \cmidrule{1-8}
            \multicolumn{3}{c|}{V2VNet} & 0.6572 & 0.6476 & 0.6336 & 0.6144 & 0.5974 \\  
            \multicolumn{3}{c|}{V2X-ViT} & 0.7202 & 0.7098 & 0.6954 & 0.6733 & 0.6614 \\  
            \multicolumn{3}{c|}{Where2comm} & 0.7103 & 0.7087 & 0.7003 & 0.6904 & 0.6793 \\  
            \multicolumn{3}{c|}{Where2comm+SyncNet} & 0.7324 & 0.7321 & 0.7238 & 0.7100 & 0.6953 \\  
            \multicolumn{3}{c|}{CoBEVFlow} & 0.7419 & 0.7397 & 0.7347 & 0.7237 & 0.7145 \\  
            \multicolumn{3}{c|}{Ours} & \textbf{0.7481} & \textbf{0.7411} & \textbf{0.7378} & \textbf{0.7289} & \textbf{0.7164} \\  \cmidrule{1-8}

            \multicolumn{3}{c|}{Model / Metric} & \multicolumn{5}{c}{AP@0.7 $\uparrow$} \\  \cmidrule{1-8}
            \multicolumn{3}{c|}{V2VNet} & 0.4982 & 0.4879 & 0.4705 & 0.4624 & 0.4512 \\  
            \multicolumn{3}{c|}{V2X-ViT} & 0.6046 & 0.5916 & 0.5758 & 0.5653 & 0.5611 \\  
            \multicolumn{3}{c|}{Where2comm} & 0.6066 & 0.6020 & 0.5875 & 0.5771 & 0.5703 \\  
            \multicolumn{3}{c|}{Where2comm+SyncNet} & 0.6246 & 0.6178 & 0.6032 & 0.5941 & 0.5877 \\  
            \multicolumn{3}{c|}{CoBEVFlow} & 0.6264 & 0.6217 & 0.6112 & 0.6042 & 0.5954 \\  
            \multicolumn{3}{c|}{Ours} & \textbf{0.6431} & \textbf{0.6366} & \textbf{0.6252} & \textbf{0.6209} & \textbf{0.6124} \\  
            \bottomrule
        \end{tabular}
    \end{adjustbox}
    \label{table2}
\end{table}

\subsection{Qualitative Results}

\begin{figure}[htbp]
  \centering
  \includegraphics[width=\columnwidth, trim=24pt 21pt 75pt 24pt, clip]{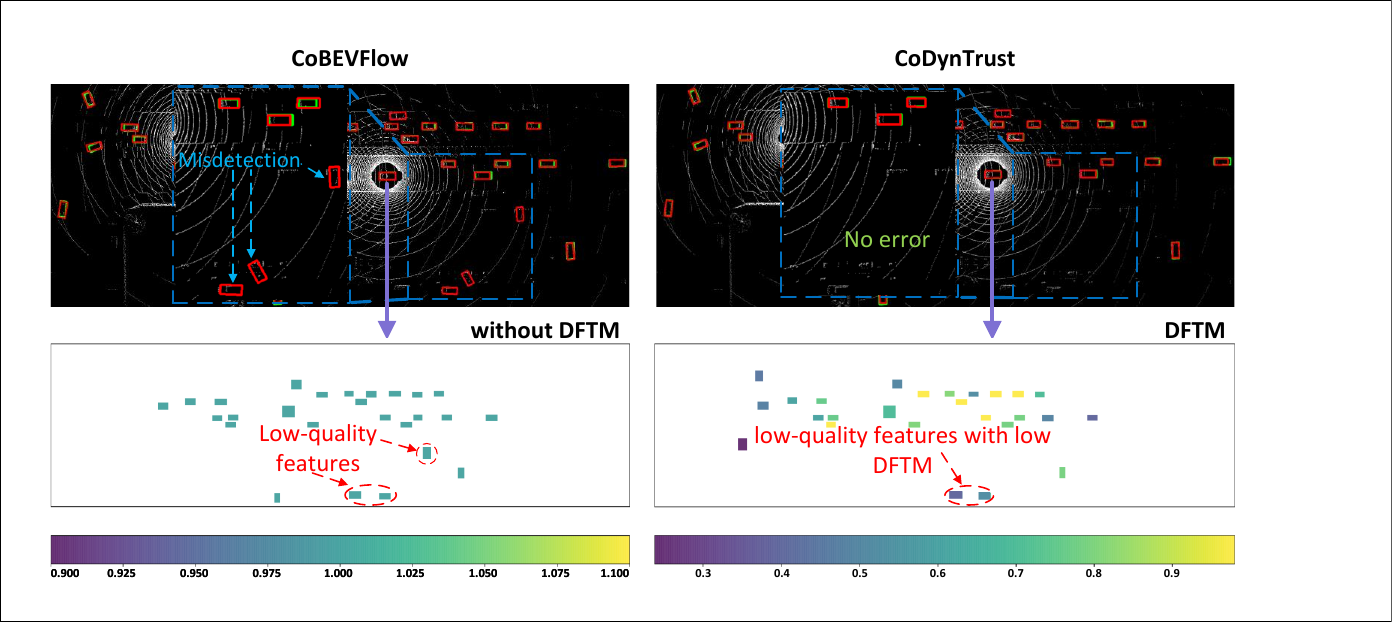}
  \caption{Visualization of CoBEVFlow and CoDynTrust detection results on V2XSet, with an expectation of a 300ms time interval. CoDynTrust shows better detection quality compared to CoBEVFlow. \textcolor{red}{Red} boxes indicate detection results, while \textcolor{green}{green} boxes represent ground truth.}
  \label{vis}
\end{figure}

Fig.~\ref{vis} shows the detection results of CoDynTrust and CoBEVFlow on V2XSet dataset, with a expected time interval of 300ms. Red boxes indicate detection results, while green boxes represent the ground truth. We compare how the two models handle features within the ROI. In complex temporal asynchrony scenarios, CoBEVFlow struggles to distinguish individual vehicle detections, leading to the sharing of low-quality detection information harms overall performance. In contrast, CoDynTrust retains reliable detections, suppressing noise and ensuring robustness.

\subsection{Ablation Studies}

We conducted ablation experiments on DAIR-V2X dataset to evaluate the effectiveness of uncertainty quantification, DFTM, and Hybrid Fusion. Table~\ref{tab:ablation} shows the results, highlighting: i) Uncertainty quantification alone provides limited performance improvement; ii) Combining uncertainty quantification with DFTM results in a significant performance boost; iii) The Hybrid Fusion mechanism, specifically designed for DFTM, further optimizes feature fusion and provides an additional performance gain.

\begin{table}[htbp]
    \centering
    \caption{Ablation study results on DAIR-V2X, \textbf{DFTM}: Dynamic Feature Trust Modulus.}
    \label{tab:ablation}
    \begin{adjustbox}{max width=\columnwidth }
        \begin{tabular}{ccc|c|c}
            \toprule
            \multicolumn{3}{c|}{Modules} & \multirow{2}{*}{AP@0.5 $\uparrow$}& \multirow{2}{*}{AP@0.7 $\uparrow$}  \\  \cmidrule{1-3}
            Uncertainty & DFTM & Hybrid Fusion & & \\  \cmidrule{1-5}
             & &  & 0.7419 & 0.6264 \\ 
            \ding{51} & & & 0.7420 & 0.6265 \\    
            \ding{51} & \ding{51} & & 0.7472 & 0.6334 \\  
            \ding{51} & \ding{51} & \ding{51} & \textbf{0.7481} & \textbf{0.6431} \\
            \bottomrule
        \end{tabular}
    \end{adjustbox}
\end{table}

\section{CONCLUSION}

We proposed CoDynTrust to tackle temporal asynchrony in collaborative perception. Using DFTM for ROI reliability and linear extrapolation for motion prediction, CoDynTrust enhances interpretability and feature fusion. Experiments show it consistently outperforms existing methods in asynchronous settings and demonstrates strong robust. Additionally, DFTM can also be propagated to downstream tasks, enhancing safety and reliability in autonomous driving. Future work will focus on optimizing DFTM generation and explore its further application in downstream modules.

\section{ACKNOWLEDGE}

This work was supported in part by the National Natural Science Foundation of China (62072321), the Six Talent Peak Project of Jiangsu Province (XYDXX-084), the Science and Technology Program of Jiangsu Province (BZ2024062), the Natural Science Foundation of the Jiangsu Higher Education Institutions of China  (22KJA520007), Suzhou Planning Project of Science and Technology (2023ss03). 

\addtolength{\textheight}{0cm}   









\bibliographystyle{IEEEtran} 
\bibliography{references,IEEEabrv} 

\end{document}